\documentclass[10pt,twocolumn,letterpaper]{article}

\usepackage{3dv}
\usepackage{times}
\usepackage{epsfig}
\usepackage{graphicx}
\usepackage{amsmath}
\usepackage{amssymb}
\usepackage{booktabs} % For formal tables
\usepackage[ruled]{algorithm2e} % For algorithms

\providecommand{\shortcite}[1]{\cite{#1}}

\usepackage[pagebackref=true,breaklinks=true,letterpaper=true,colorlinks,bookmarks=false]{hyperref}

\threedvfinalcopy % *** Uncomment this line for the final submission

\def\threedvPaperID{35} % *** Enter the 3DV Paper ID here

\ifthreedvfinal\fi

\usepackage{color}
\usepackage{multirow}
\usepackage{rotating}
\usepackage{amsmath}
\usepackage{amsfonts}
\usepackage{dsfont}
\usepackage{array}
\usepackage{wrapfig}
\usepackage{subcaption}
\usepackage{comment}
\definecolor{turquoise}{cmyk}{0.65,0,0.1,0.1}
\definecolor{purple}{rgb}{0.65,0,0.65}
\definecolor{darkgreen}{rgb}{0.0, 0.5, 0.0}
\definecolor{darkred}{rgb}{0.5, 0.0, 0.0}
\definecolor{darkblue}{rgb}{0.0, 0.0, 0.5}
\definecolor{blue}{rgb}{0.0, 0.0, 1.0}

\newcommand{\setR}{\mathds{R}}

\begin{document}

%%%%%%%%% TITLE
\title{Parsing Geometry Using Structure-Aware Shape Templates}

%\author{ Vignesh Ganapathi-Subramanian \\
%\and
%% omit the following lines up until the closing ``''.
%% Additional authors and addresses can be added with ``\and'',
%% just like the second author.
%% To save space, use either the email address or home page, not both
%Olga Diamanti \\
%\and
%Soeren Pirk \\
%\and
%Chengcheng Tang \\
%\and
%Matthias Nie\ss ner \\
%\and
%Leonidas J. Guibas \\
%}

\author{ Vignesh Ganapathi-Subramanian\\
\small Stanford University\\
{\tt\small vigansub@stanford.edu}
% omit the following lines up until the closing ``''.
\and
Olga Diamanti\\
\small Autodesk Inc., Stanford University\\
{\tt\small diamanti@stanford.edu}
\and
Soeren Pirk \\
\small Stanford University\\
{\tt\small pirk@google.com}
\and
Chengcheng Tang \\
\small Stanford University\\
{\tt\small tangcc@stanford.edu}
\and
Matthias Nie\ss ner \\
\small TU Munich, Stanford University \\
{\tt\small niessner@tum.de}
\and
Leonidas J. Guibas \\
\small Stanford University \\
{\tt\small guibas@cs.stanford.edu}
}

\maketitle
% \thispagestyle{empty}

%%%%%%%%% ABSTRACT
\begin{abstract}
Real-life man-made objects often exhibit strong and easily-identifiable structure, as a direct result of their design or their intended functionality. Structure typically appears in the form of individual parts and their arrangement. Knowing about object structure can be an important cue for object recognition and scene understanding - a key goal for various AR and robotics applications. However, commodity RGB-D sensors used in these scenarios only produce raw, unorganized point clouds, without structural information about the captured scene. Moreover, the generated data is commonly partial and susceptible to artifacts and noise, which makes inferring the structure of scanned objects challenging. In this paper, we organize large shape collections into parameterized shape templates to capture the underlying structure of the objects. The templates allow us to transfer the structural information onto new objects and incomplete scans. We employ a deep neural network that matches the partial scan with one of the shape templates, then match and fit it to complete and detailed models from the collection. This allows us to faithfully label its parts and to guide the reconstruction of the scanned object. We showcase the effectiveness of our method by comparing it to other state-of-the-art approaches.
\end{abstract}

%%%%%%%%% BODY TEXT

\maketitle

\section{Introduction}

% \olga{this is just the rough idea}

In all their variability, real-life man-made objects are often designed with certain basic principles in mind, relating to their target functionality, affordances, physical and material constraints, or even aesthetics.
%All of these parameters affect the choice of structural layouts for an object.
As a result, most objects can be described by common structural forms or patterns. For example, chairs often follow a model of ``four-legs'', with/without armrests, ``S-''shaped, or swivel. Object structure can be modeled by identifying the most common part layouts, spatial interrelations and part symmetries.

\begin{figure}[t!]
  \includegraphics[width=\columnwidth]{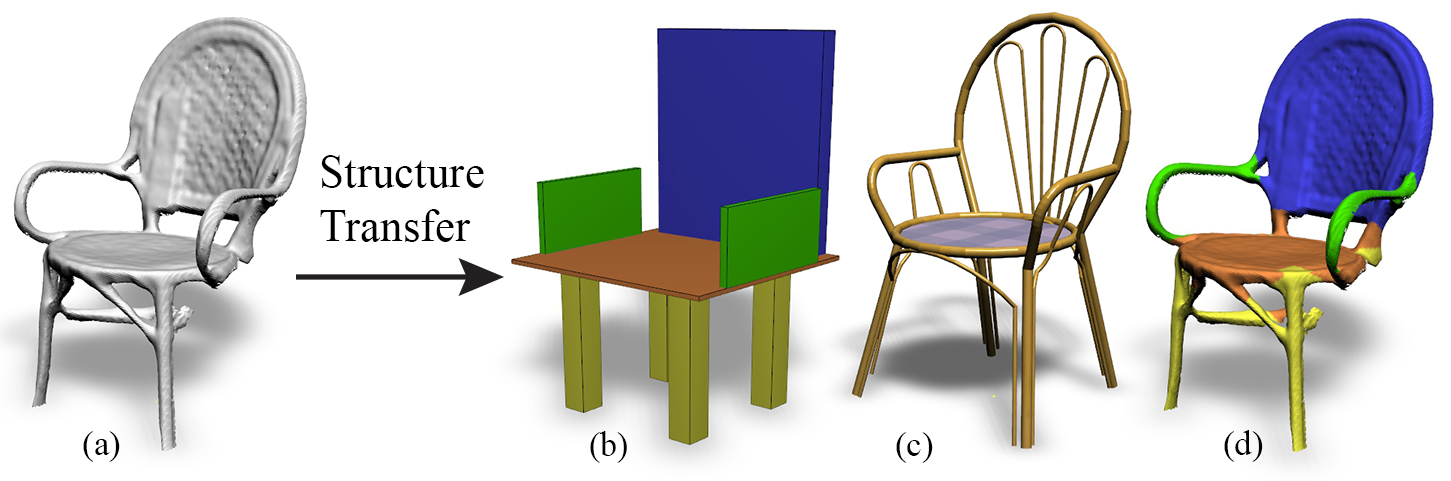}
  \centering
  \caption{ \small  Structure transfer with templates. Given a partial scan (a), we detect and fit a matching structural template (b). A similar complete shape from a collection, recovered and deformed to fit the scan (c). (d) Semantic part labels transferred onto the scan.}
  \label{fig:teaser}
  \vspace{-8mm}
\end{figure}

Associating an unknown object with a particular structural pattern can help with understanding the object's properties and purpose. In particular, structure can help in cases where part of the object geometry is missing, due to the object being only partially observed. This is a common scenario in 3D sensing pipelines, e.g. those arising in augmented reality (AR) and robotics, or used for data acquisition for graphics/visual effects applications, where a scene is reconstructed for further editing. 3D sensing pipelines typically rely on scanning a scene through RGB-D sensors, which commonly produce incomplete, noisy, and unorganized point scans. The resulting reconstructed geometry often is difficult to parse and may fail to meet the requirements of applications that use more precise representations of shapes, e.g. CAD models. Faithfully inferring object structure and using it to recover a corresponding complete surface mesh from partial sensor data remains one of the fundamental problems in shape reconstruction research.
%
%
% Associating an unknown object with a particular structural pattern can help with understanding the object's properties and purpose, especially when the object is only partially observed. This association can be immensely helpful in 3D sensing pipelines, in particular those arising in augmented reality (AR) and robotics, as well as in graphics-oriented applications, such as structure-aware reconstruction and design. These tasks typically rely on sensing a particular scene through geometry scanning devices, e.g., RGB-D scanners or cameras, which can only produce raw geometry in the form of incomplete, noisy, and unorganized point clouds. Inferring object structure in these scenarios is a challenging task, but it can also bring great benefit, by helping infer the missing geometry and better understand the shape semantics. In this sense, structure inference and geometry parsing are closely connected in this setting. The focus of this work is to obtain generalizable knowledge for inferring structure from partial geometries of unknown shapes.

 %Inferring object structure in these scenarios is a challenging task, but it can also bring great benefit, by helping infer the missing geometry and better understand the shape semantics. In this sense, structure inference and geometry parsing are closely connected in this setting. The focus of this work is to obtain generalizable knowledge for inferring structure from partial geometries of unknown shapes.

We introduce a pipeline for transferring structural information from large shape collections onto unknown scanned shapes. We model object structure via a small number of hand-crafted templates, to capture the structural patterns exhibited by shapes of a particular category. Our hand-crafted templates are designed to represent an abstraction of the shape. This is similar to how designers model shapes, often starting with a rough structure in mind and based on the desired utility -- only afterwards are the details carved out. Manually providing the templates is meant to emulate this process. Our templates are abstract, lightweight, yet meaningful enough to account for thousands of shapes in various forms. Each template consists of axis-aligned boxes in a specific spatial arrangement, providing knowledge about the structural organization of the shape and the relationship of its parts. We parameterize the templates so as to fit them, via state-of-the-art optimization, to a particular shape.

% In conjunction with structural information, we encode geometric and stylistic similarities between shapes by building a common shape space for all shapes across categories.
% The combination of structural information (encoded by the templates) and geometric similarity (encoded by the shape space) constitutes a representation for inferring structure from partial geometries obtained through scanning.
We then leverage the structural information encoded by the templates for the entire shape collection to learn their structure. This allows us to identify partial shapes obtained through scanning. To address this problem of inferring the structure of a shape from its partial geometry, we employ a deep neural network trained on partial views of shapes. The network is able to detect shape structure. Using this, one can identify the scanned object by retrieving the closest shape from the collection and fitting it to the scan. Additional applications of the templates include part labels for partial scans, and recovery of a fully meshed CAD model to replace the scan in scene modeling scenarios (Fig.~\ref{fig:teaser}).

\section{Related Work}
%Representing and understanding shapes has been an area of research in computer graphics and geometric modeling for many decades.
Due to the complexity and variability of objects, efforts on representing and understanding shapes focus on their
%fundamental methods to represent and describe the repetitive geometric nature of shapes~\cite{rs71012763}, to more principled ways
%for the automatic generation of specific object classes, such as trees, buildings, or entire landscapes~\cite{10.1111:cgf.12276}. More recent methods not only model shapes and their variations, but also explore more nuanced ways
 structure~\cite{Mitra14}, their features and similarities~\cite{CGF:CGF12734}, the semantic meaning of individual parts~\cite{Xu16}, and even the creative process of shape modeling~\cite{Cohen-Or:2016:IMC:2893030.2893075}.

\textbf{Structure-aware Representations.} It has been recognized that the structural organization of shapes plays an eminent role in modeling and reconstructing shapes. Existing  approaches focus on identifying shape parameters~\cite{Gal:2009:IAA:1531326.1531339}, relations~\cite{Kim:2012:ECM:2185520.2185550,Kalogerakis:2012:PMC:2185520.2185551}, symmetries~\cite{Pauly:2008:DSR:1360612.1360642,Xu:2012:MPI:2366145.2366200}, correspondences~\cite{Ovsjanikov:2012}, combinatorial variations~\cite{Funkhouser:2004,Bokeloh:2011:PSD:2070781.2024157}, or co-variations~\cite{vanKaick:2013:CAS:2461912.2461924}.
%or on directly decomposing shapes~\cite{CGF:CGF1103,Huang:2011:JSS:2070781.2024159}.
%, also while considering task-specific constraints, such as saving material in fabrication processes~\cite{Hu:2014:APS:2661229.2661244}
%Several methods exist to infer the structure of shapes through consistent segmentation and labeling of shape parts.
%Kalogerakis et al.~\shortcite{Kalogerakis:2010} and Wang et al.~\shortcite{Wang:2012:ACS:2366145.2366184} introduce data-driven and semi-supervised pipelines for inferring semantic part labels from unstructured meshes.
More recently, Schulz et al.~\shortcite{Schulz:2017} show that discrete and continuous shape features can be represented as low-dimensional manifolds and then used to retrieve individual shapes from large shape collections. All these techniques identify and represent shape structure, however, they do not aim at reconstructing or replacing partial scans.

\textbf{Shape Templates and Part-based Models.} Shape templates have proven to be an effective tool for inferring higher-level knowledge of shapes, not only in the context of 3D geometry~\cite{Ovsjanikov:2011}, but also for various image processing tasks ~\cite{Felzenszwalb:2010:ODD:1850486.1850574, NIPS2012_0057, 6130367}.
%image processing tasks, such as object recognition~\cite{Felzenszwalb:2010:ODD:1850486.1850574}, segmentation~\cite{NIPS2012_0057}, and pose estimation~\cite{6130367}.
Kim et al.~\shortcite{Kim:2012:AIE:2366145.2366157} employ part-based templates to encode deformations and fit object labels to scanned point clouds. Kalogerakis et al.~\shortcite{Kalogerakis:2012:PMC:2185520.2185551} learn distributions of parts to encode part placements, Kim et al.~\shortcite{Kim:2013:LPT:2461912.2461933} propose a fully automatic approach for inferring probabilistic part-based templates to capture style and variation of shapes in large model collections. Unlike the existing approaches, we use shape templates to organize large shape repositories and to learn the structure of shapes and their parts.

\textbf{Shape Reconstruction.} An especially difficult problem is the reconstruction of shapes from unstructured or incomplete data, such as partial meshes or point clouds. Symmetry-driven reconstruction  leverages symmetric properties in the scanned data to complete occluded regions~\cite{Pauly:2008:DSR:1360612.1360642,Bokeloh:2010:CPS:1778765.1778841}. Sipiran et al.~\shortcite{CGF:CGF12481} rely on local features and fit surface functions to incomplete point clouds. Shen et al.~\shortcite{Shen:2012:SRP:2366145.2366199} propose a bottom up structure recovery pipeline that allows to replace incomplete point scans of man-made objects by aligning them to existing shapes in a repository.%Similar to our approach,
Sung et al.~\shortcite{Sung:2015:DSP:2816795.2818094} use 3D shape collections and exploit symmetry relations to infer global structure and complete point clouds with substantial occlusion. However, they concentrate on learning the distributions of shape part features, and use the collection to retrieve only individual shape parts, not entire shapes to match the partial point cloud. %While we are also able to replace partial scans with repository models, we also exploit the inferred structure to directly densify the actual scan.
\begin{figure}[t]
  \begin{centering}
  \includegraphics[width=1.0\linewidth]{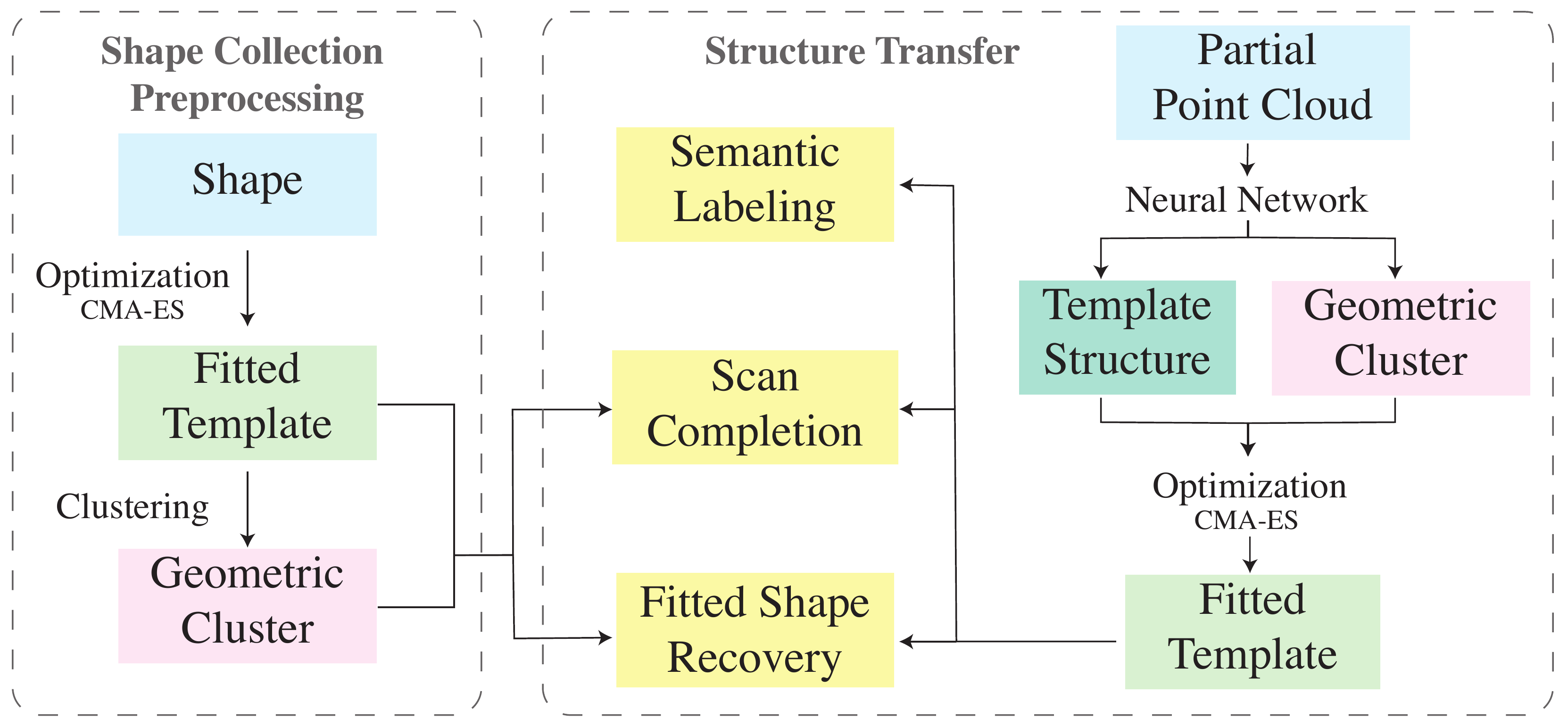}

  \caption{ \small  Overview: we propose a two-stage framework for parsing shapes with the end goal of retrieving structure from partial point clouds. In the first stage (shape collection preprocessing - left), we organize the shapes in our collection according to their structure, using templates. In the second stage (structure transfer to partial shapes - right), we train a deep neural network to retrieve a proper geometry-based cluster and a suitable structure template. We further refine the template parameters through optimization.}
  \label{fig:overview}
  \vspace{-4mm}
  \end{centering}
\end{figure}

\section{Overview}
\label{sec:overview}

Our framework works in two stages (Fig.~\ref{fig:overview}).
% First, we organize the shapes in a shape collection in terms of their structural patterns and approximate dimensions. Second, we apply the obtained knowledge to infer the structure of unseen shapes represented as partial and unstructured point scans.
%
In the first step, we fit a set of pre-defined shape templates to shapes of a shape collection. This is necessary, as shape repositories commonly do not provide information about the structure of shapes, but instead only store surface meshes of the models. Fitting templates allows us to infer the structural organization of the shape collection. % Namely, at this first stage, each shape in the collection is assigned to one of these structure-aware templates, which is then tailored to fit that individual shape.
Moreover, we cluster shapes according to their template parameterization. This provides a geometric organization of the shapes, which can be leveraged to perform approximate shape retrieval from the collection.
In the second step, we employ the template organization imposed on the collection above to infer shape structure for unseen shapes with partial geometry. We use the inferred structure to retrieve and fit known shapes from the database to the partial scan, which can directly be used as a proxy shape for scanned geometry. Additionally, the structure allows to identify and annotate shape parts and to reconstruct the object in a structure-aware manner.
%Annotation of shape parts and densification of the partial scan are also aided by inferring the structure.
% comparisons are not a part of the overview, they are evaluation:)
% Finally, we compare our methods with state-of-the-art methods to show that the knowledge of structure learned and transferred based on organized shape database is essential in understanding the structure of partial scans and parsing the geometry in the wild.

%!TEX root = ../egpaper_submission.tex

%\section{Imposing structure onto a collection of shapes}
\section{Shape Collection Preprocessing}
\label{sec:collection_structure}

In this section we describe the first stage of our pipeline, which involves organizing the shapes in the collection $\mathcal S$ using our primitive-based shape templates. They provide a structure-aware shape representation. We assume that all the models in $\mathcal S$ are pre-aligned, which makes defining a common frame of reference possible.

\subsection{Structure-Aware Templates}
We first define a set of structural templates.
% Each template is specific to a particular shape family (e.g., chairs), and captures a particular structural ``mode'' frequently found among shapes in that family.
Each template captures a particular structural ``mode'' frequently found among shapes in that family -- in the ``chair'' example, one template might capture the common four-legged pattern, while another the swivel structure.%Note that a structural template can be applicable to more than one shape family -- e.g. sofas and chairs can both exhibit four-legged patterns, encoded by a single template.
We fit these templates to all shapes of that family in the collection, and choose the most appropriate template for each shape.

\paragraph*{Template Definition.}
A \emph{template} is a collection of deformable axis-aligned box primitives and connectors, forming a tree graph structure. Each box  represents a structural part of a shape.
%Additional meta-information to capture higher-level structural relations and symmetries.
Specifically, a template consists of:

{\bf Nodes.}  Graph nodes (vertices) are associated with the box primitives in the template. Each box is described by 6 \emph{parameters}, namely its position and size in 3D.

{\bf Connectors.} Template graph edges ``glue'' together adjacent boxes, structurally attaching them and constraining their dimensional parameters. Connectors also define the relative positions of child boxes to parent boxes. We only model attachment at right angles.

{\bf Groups.} Symmetric relationships are often present among structural parts of shapes, e.g. between the four legs of a four-legged chair. In terms of our templates, symmetries are modeled by symmetric structural constraints between graph nodes, which require that the parts (in this example, the four leg boxes) are identical. This symmetry meta-information is encoded via groups in the graph.

We pre-design a set of $N_T = 21$ templates in total. Each shape family is associated with a subset of these templates, but a single template can be shared by more than one shape family. For example, tables and desks often have similar structure, which is reflected by them sharing one of their templates. The graph structure of the templates stays fixed throughout our pipeline, but we can tune a template to a particular shape by finding an optimal configuration of template box parameters. The fitted template is then a structure-aware representation of the shape, with semantic information about the shape parts encoded as meta-information.

\begin{figure}[t]
  \includegraphics[width=\linewidth]{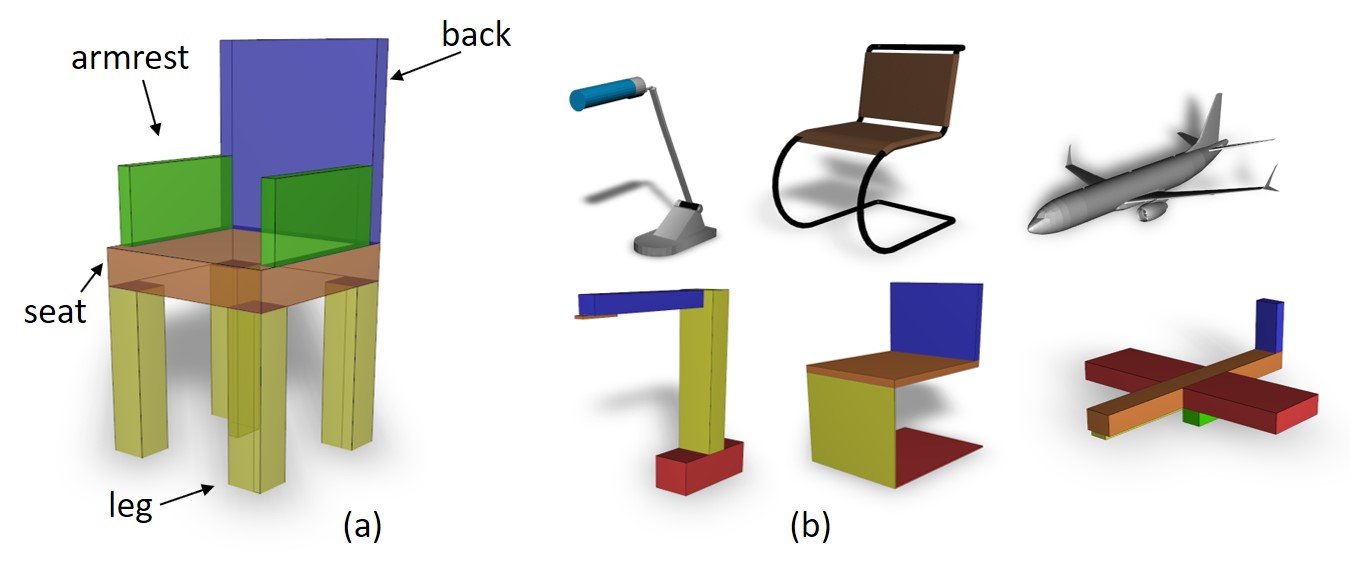}
  \caption{ \small (a) A shape template with semantic information embedded in its parts (b) Top - various shapes from the collection: a lamp, a chair with folding legs, and an airplane. Bottom - The fitted templates, with parameters that best approximate those shapes.}
  \label{fig:template_correspondences}
\vspace{-5mm}
\end{figure}

\subsection{Template Selection and Fitting}
\label{sec:template_selection}
We are now given a shape $S$ of a particular family, in the form of an unstructured point cloud sampled from a database shape. The aim is to find which template structure $T = \mathcal T(S)$  best approximates the structure of $S$, and compute values for its  parameters  $\mathcal P(S, T)$ (box sizes and locations) to fit the shape's geometry. We proceed in two stages: first, we fit all templates pertaining to the shape's family  to the shape, e.g. chair templates to a chair shape. Then, we select the best-fitting template from that set.

\subsubsection{Template Fitting}
\label{sec:template_fitting}

We aim to fit a template structure $T$ to an input point cloud $S$. The template consists of boxes $B_i$, $i = 1, 2, ... , N$.
For simplicity, in the following discussion we denote both the template and its shape-dependent parameters $\mathcal P(S, T)$ as $T$.
The optimal parameter values are found by solving an optimization problem
% \begin{displaymath}
$
  T^* = \mathop{argmin}\limits_T E_{\text{total}} (T) = \sum_{i} \lambda_iE_i(T)
$
% \end{displaymath}
where $i$ ranges over individual energy terms.
%\begin{eqnarray}
%E_{\text{total}} (T) &=& \lambda_{\text{proj}}E_{\text{proj}} (T)+ \lambda_{\text{bbox}}E_{\text{bbox}} (T)\nonumber\\
%&&+ \lambda_{\text{min}}E_{\text{min}} (T)+ \lambda_{\text{dis-ent}}E_{\text{dis-ent}}(T)
%\label{equ:optimization_objective}
%\end{eqnarray}
The various terms in $E_{\text{total}}$ encourage a close match between the ``box-like'' template shape and the input point cloud at the optimum. We used $\lambda_{\text{proj}} = 0.3$, $\lambda_{\text{bbox}} = 1$,  $\lambda_{\text{min}} = 0.8$,  $\lambda_{\text{dis-ent}} = 0.4$ in all our experiments. The individual energy terms are detailed below. A qualitative evaluation of various energy terms is discussed in the supplementary material.

{\bf Projection:} The sum of distances from all points in the point cloud $S$ to the template geometry: $E_{\text{proj}}(T)  = \sum_{\mathbf{p} \in S} \text{min}_{j = 1, 2, ... , N} d(\mathbf{p}, B_j)$,
%\begin{equation}
%E_{\text{proj}}(T)  = \sum_{\mathbf{p} \in S} \text{min}_{j = 1, 2, ... , N} d(\mathbf{p}, B_j).
%\end{equation}
where $d(\mathbf{p}, B_j)$ is the minimum projection distance from point $\mathbf{p}\in\setR ^3$ to box $B_j$ in the template. The projection term ensures that the optimization produces a well-fitting template.

{\bf Bounding Box:} The difference in size  (3D volumes) between the bounding box of the point cloud $S$ and that of the template $T$: $E_{\text{bbox}}(T) = |Vol(T_\text{bbox})-Vol(S_\text{bbox})|$
%\begin{equation}
%E_{\text{bbox}}(T) = |Vol(T_\text{bbox})-Vol(S_\text{bbox})|
%\end{equation}

{\bf Minimalism:} The total volume of the template in space: $E_{\text{min}}(T)  = \sum_{B_i \in T} Vol(B_i)$
%\begin{equation}
%E_{\text{min}}(T)  = \sum_{B_i \in T} Vol(B_i)
%\end{equation}
With this term, the thinnest possible set of boxes is fitted to the point cloud, ensuring that the template geometry is no bigger than need be. While this term overlaps with the bounding-box energy, we found that it promotes faster optimization convergence.

{\bf Disentanglement:} The amount of overlap between boxes: $E_{\text{dis-ent}}(T)  = \sum_{B_i \in T} \sum_{B_j \in T, B_j \neq B_i} Vol(B_i \cap B_j)$
%\begin{equation}
%E_{\text{dis-ent}}(T)  = \sum_{B_i \in T} \sum_{B_j \in T, B_j \neq B_i} Vol(B_i \cap B_j)
%\end{equation}
This term requires that template boxes don't spatially obstruct each other, since they are meant to capture distinct semantic shape parts.

\subsubsection{Optimization and Template Selection}
\label{sec:optimization}
The energy $E_\text{total}(T)$ %in Eq.~\eqref{equ:optimization_objective}
is highly non-convex, requiring an optimization scheme that avoids local minima. Evolutionary optimization strategies are particularly appropriate for this task \cite{hansen1996_adapting-arbitrary-normal-mutation}, since they continuously adapt the step size based on previous observations and utilize a randomization factor to probe large energy landscapes. In addition, they are more easily parallelizable compared to classical gradient-based techniques.  We employ the Covariance Matrix Analysis - Evolutionary Strategy (CMA-ES)~\cite{hansen2006cma}, which uses correlations between different variables to determine the step size.

Given the non-convexity and in order to aid convergence, we initialize the optimization based on solutions of previously successful optimizations for other shapes of the same family. We eventually choose the optimal set of parameter values across all runs as the best fit for a given template.
Once all candidate templates have been fitted to a given shape, we select the best-fitting one $T^*$ as the one that minimizes $E_\text{total}(T)$ over all candidate templates.
%Since the energy functional has been designed exactly to measure how well a given template fits the shape, we select as best-fitting the one for which the value of the energy functional at the optimum of parameter configurations is minimal.
A caveat to performing template selection when two similar templates are used and also various convergence timings are discussed in the supplementary material.

\subsection{Structural Shape Clustering}
\label{sec:shape_clustering}

We further use the templates to cluster the shapes in the collection into groups of similar structure and rough dimensions. We first group the shapes in the collection according to their best fitting template structure. Then, we use the vectors of template parameters (box dimensions) to further divide the shapes of each group into clusters of shapes of similar part dimensions via k-means. We use 10 clusters per template ($N_C=210$ clusters total). Each cluster is then associated with a specific template, but can contain shapes from different families, since the same template can be shared by more than one family (e.g. tables and desks). The clusters will  be used to inform the structure identification of partially scanned shapes.

\section{Structure Transfer to Partial Shapes}
\label{sec:structure_transfer}

Having used the templates to organize the data in our shape collection in terms of their structure, we now transfer structure to partial scans of objects not present in the shape collection. We use the structural information to better understand the partial object's shape, and structure it by assigning and fitting it to one of the known structural patterns found in the collection.

\subsection{Inferring Structure from Partial Scans}
\label{subsec:partial_shape_structure_inference}

The input to the structure inference stage is a partial point scan $S_{\text{partial}}$. The output is one of the  templates, fitted to $S_{\text{partial}}$; this process imposes structure on the partial point cloud. %The na\"ive approach of testing the scan against all possible templates for all shape families would be prohibitively slow.
We train a deep neural network that (indirectly) assigns the partial shape to a particular shape template; it predicts structure from partial geometry. Trained with simulated partial views, the network learns to ignore various artifacts and noise commonly found in real RGB-D scans. Once the network selects the template pattern, we further optimize template parameters to fit the scan ( Section~\ref{sec:template_fitting}).

Note that the template assignment, on its own, is not sufficient to identify a shape, since a particular template graph can be shared by multiple shape families. Not knowing the shape family has the additional negative side-effect that we cannot intelligently initialize the fitting CMA-ES optimization as described in Section~\ref{sec:optimization}, which can adversely affect performance both in terms of runtime and quality of results. Thus, instead of having the network directly predict the template structure, we train it to predict the \emph{shape cluster} instead; since each cluster is associated with a particular template structure, the network also indirectly predicts the template. The predicted cluster also provides, via its shapes, a set of candidate template fits for that particular template pattern, which can  be used as initialization to get the optimal fit of the template pattern to the partial scan. Thus, the network predicts, from raw partial geometry, both a rough structure pattern and part dimensions for the partial scan.

% One way to infer the shape family is to examine the partial scan's ``nearest neighbors'' in the shape collection, which  can achieved by assigning the scan to one of the structural clusters defined in Section~\ref{sec:shape_clustering}. We pose this problem of ``positioning'' the partial scan inside the collection as a machine learning problem and adapt the same network to also detect the \emph{cluster} of shapes in which the partial scan belongs \olga{why do we need both cluster and template? can they correspond to different structures? that depends on how we do the clustering}. Knowing the cluster enables inferring possible shape families the scan might belong to, since as we saw clusters and families correlate\olga{did not see this - does that property hold?}; on the other hand, since the clustering was done on all collection shapes simultaneously \olga{verify}, we are able to identify the object across categories, which can give additional flexibility in practise.
% Knowing the template and the cluster fully determines the family and structure of the partial shape. Note that the network is able to efficiently ``identify'' the object (in terms of its family, coarse structure, and approximate geometry) by operating across all collection shapes simultaneously,, thus avoiding the need for category-specific learning, which can be less time-efficient and arbitrary.

\begin{figure}[t]
  \includegraphics[width=\linewidth]{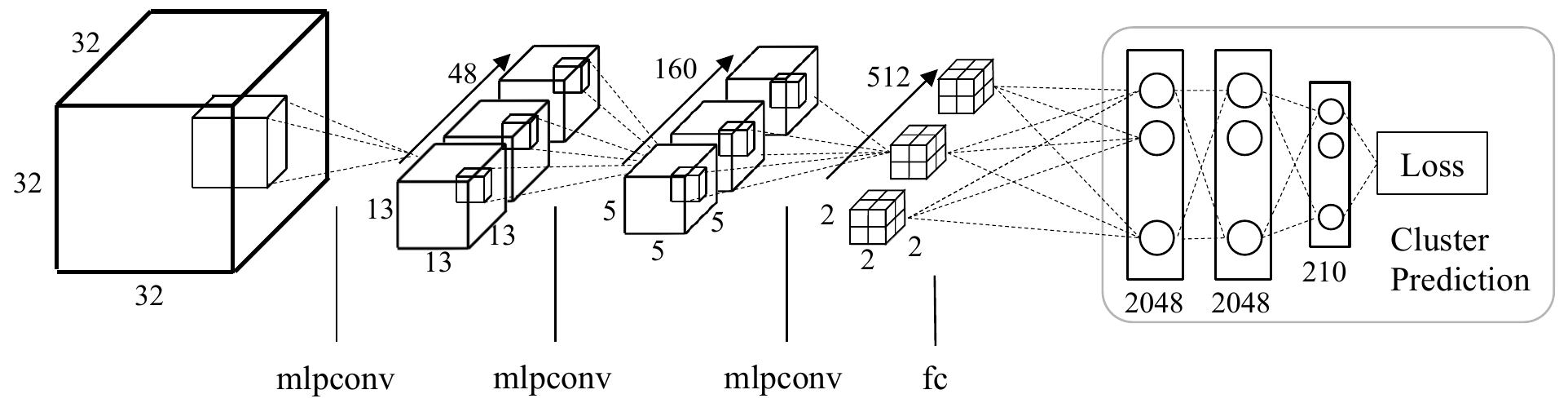}
  \vspace{-4mm}
  \caption{ \small Classification Network, inspired by \protect\cite{DBLP:journals/corr/QiSNDYG16}. Multiple mlpconv layers, which are powerful local feature extractors, are followed by a set of fully connected layers. This network learns and shape cluster from the input signed distance field of a partial point scan, and from that, the best structural template.}
  \label{fig:3dnin}
\end{figure}
\begin{figure}[t]
  \includegraphics[width=\linewidth]{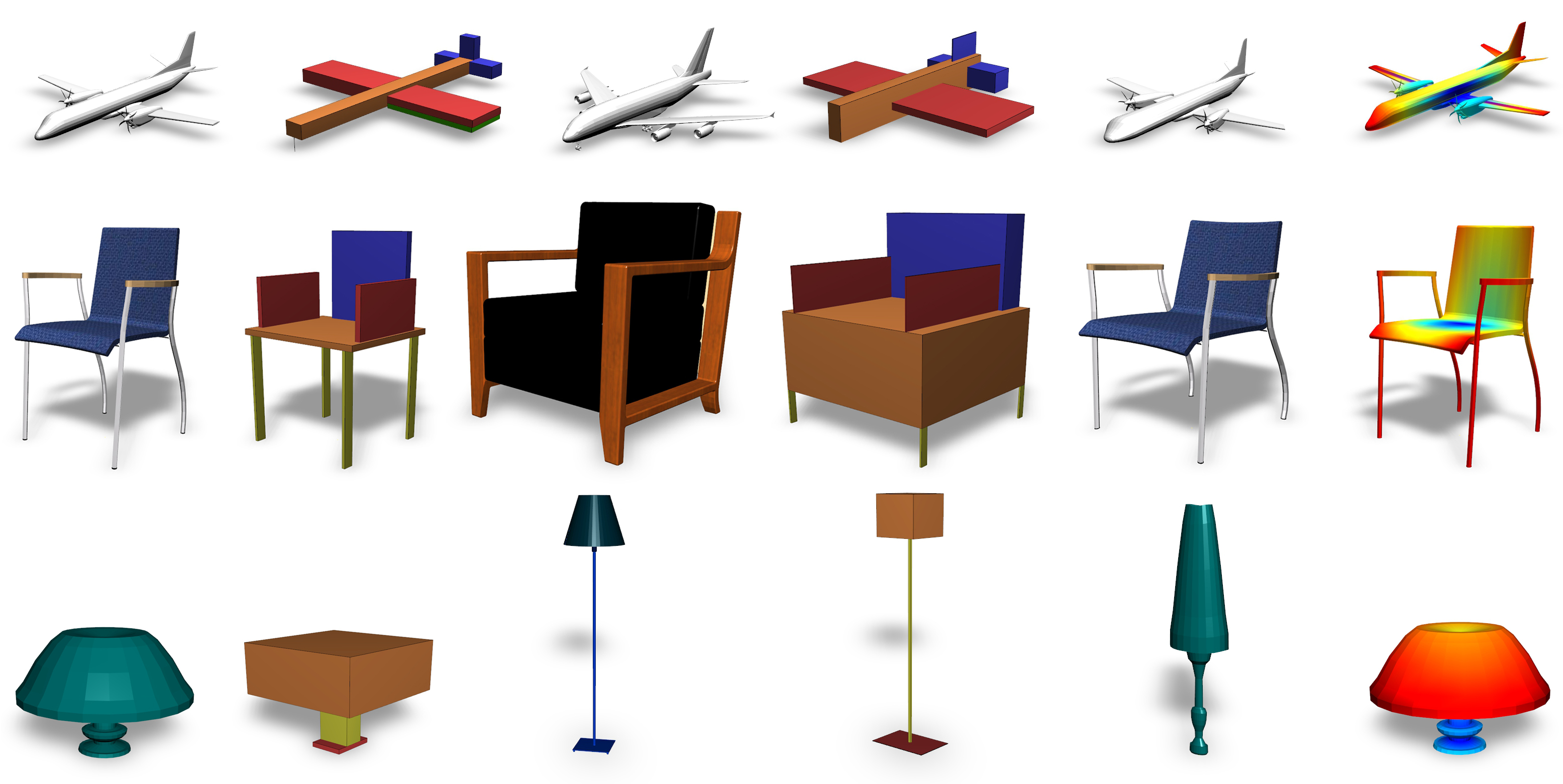}
  \caption{ \small Structure-aware shape manipulation. Left to right: source shape and fitted template, target shape and fitted template, deformed version of the source shape to match the target. The rightmost heat map indicates the normalized per-vertex magnitude of deformation induced on the source shape.}
  \label{fig:skinning}
\end{figure}

% In the following, we describe in detail the network learning pipeline, and show how it can be used for the applications of shape recovery and completion.
\subsubsection{Classification Network}
\label{subsubsec:net}
The input to the network is the signed distance field $d_{\text{partial}}$ of the partial point scan $S_{\text{partial}}$. The output is a vector of length $N_C$, encoding the probabilities with which each shape cluster corresponds to the scan.
% There are a one caveats to be noted here. It is evident that stages one and two need to be performed before stage three, since the outputs of previous two stages, act as ground-truth data to train the neural network in stage three. More specifically, the training for the first leg of the dual-classification network $\mathcal{N}$ is done on best template $t(S_i)$ for input $d_i$ (signed distance fields generated on partialization of shapes $S_i$). The training for the second leg of $\mathcal{N}$ is done on the clusters generated in Stage Two. Nevertheless, stage one and two can be performed independently and in parallel since the results of the two blocks do not influence one another.
%
The network architecture (Fig.~\ref{fig:3dnin}) uses mlpconv layers\cite{DBLP:journals/corr/QiSNDYG16}, shown to be powerful local patch feature extractors. They are followed by a number of fully connected layers and dropout layers of factor 0.2. We use cross-entropy \cite{goodfellow2016_deep-learning} as the loss function and perform data augmentation \cite{DBLP:journals/corr/QiSNDYG16} to reduce overfitting.

\subsubsection{Partial Shape Identification}
\label{subsubsec:partial_shape_identification}
The most likely shape clusters, as predicted by the network, indicate the most likely template structures for the partial point cloud $S_{\text{partial}}$. We fit templates corresponding to the $k=3$ most likely clusters to the partial shape (Section~\ref{sec:template_fitting}) and pick the best-fitting template $T_{\text{partial}}$ based on fitting error. This fully structures the partial shape and identifies its family. We initialize the optimization for each template by averaging the best-fitting parameter values for the shapes in the cluster where the template came from. The optimization also produces the optimal template parameters $\mathcal P (S_{\text{partial}}, T_{\text{partial}})$  aligning the template to partial geometry.

% With a selected template and cluster information, we . We perform this on the top three cluster and template pairs and select the best template, cluster and template parameterization based on the optimization objective.
% The optimization block that follows, uses the cluster and template values as input, and uses them for the parameter initialization and obtains the template parameterization $p(S_\text{new},t_\text{new})$ for the partial shape.

\subsection{Structure-Aware Partial Shape Recovery}
\label{subsec:partial_shape_recovery}
The identification of a partial shape, via its fitted template and family, enables \emph{shape recovery}: we can retrieve, among all shapes in the collection,  one that fits the partial point cloud. This is useful for scene understanding scenarios, in AR or robotics, where a partial point cloud can be mapped to (and possibly replaced by) an already known shape, or for editing a scanned scene for CG applications.
Since we cannot directly use shape geometry to detect the most similar shape as our input is partial, we use fitted templates to provide a rough proxy for geometric and structural similarity, via their box dimensions and locations.
% In addition, structural and geometric similarity are often correlated, as shown in Fig.s~\ref{fig:similarities} and \ref{fig:tsne_comparison}.
We thus look for a shape in the collection that matches $S_{\text{partial}}$ in terms of its parametric template fit.
We search in the $k=3$ most likely clusters as provided by the  network.
The output of this stage is a shape from the  collection $S_{\text{source}}$, along with the optimal parameter values $\mathcal P(S_{\text{source}}, T_{\text{partial}})$ that fit $S_{\text{source}}$ to the template of the partial shape, $T _{\text{partial}}$.
\vspace{-4mm}

\begin{figure}[t]
  \includegraphics[width=\linewidth]{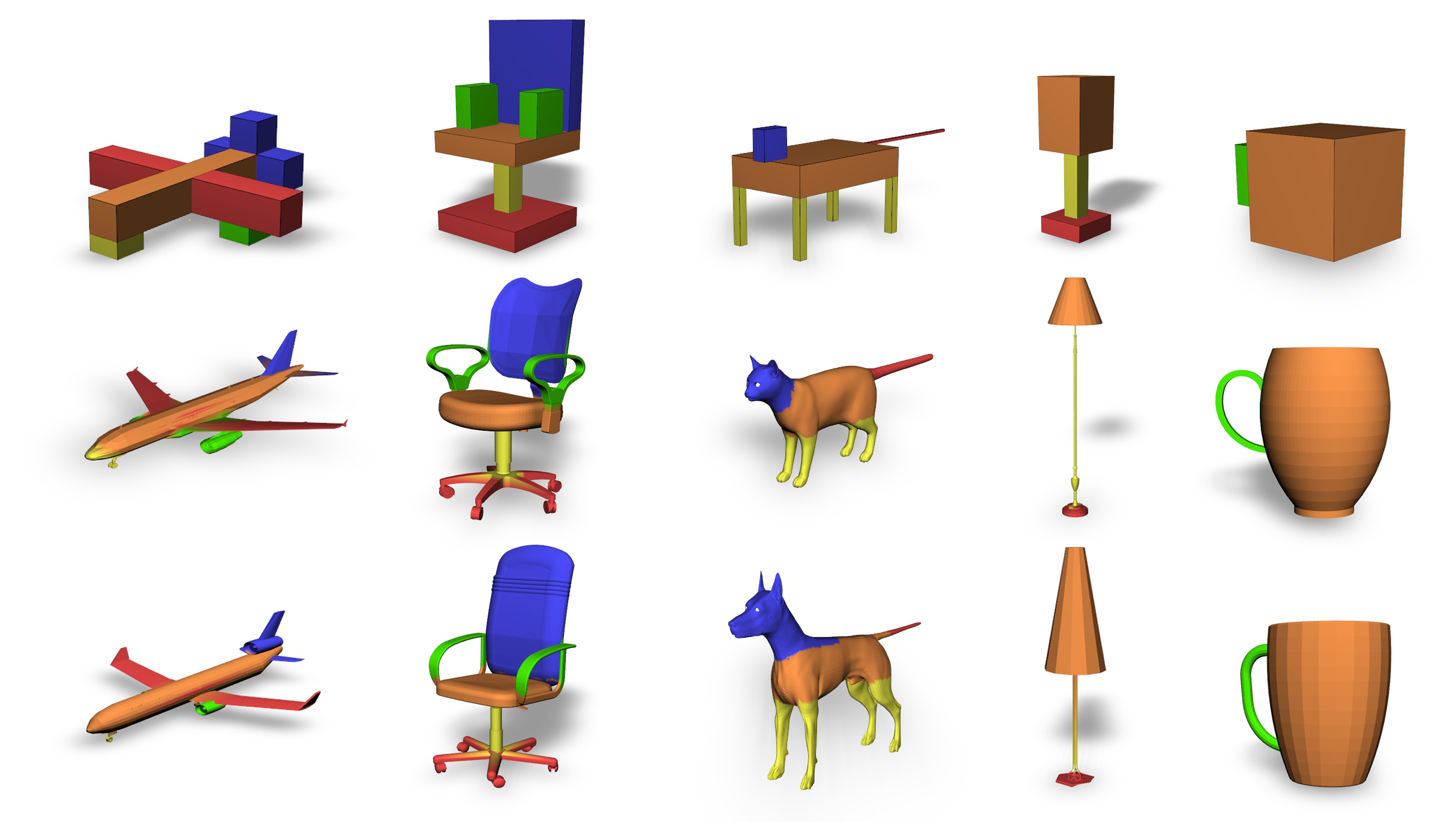}
  \caption{ \small Semantic labeling for airplanes, chairs, animals, lamps and mugs, as produced by the box-template parameters. The parts are color-coded based on their tags: \emph{airplanes}-\{(orange, body), (red, wings), (blue, tail), (green, engine), (yellow, front wheel)\}, \emph{chairs}-\{(orange, seat), (red, base), (yellow, swivel), (green, armrests), (blue, backrest)\}, \emph{animals}-\{(blue, head), (orange, torso), (red, tail), (yellow, legs)\}, \emph{lamps}-\{(orange, bulb), (yellow, stem), (red, base)\}, \emph{mugs}-\{(orange, cup),(green, handle)\}.}
  \label{fig:shape_labeling}
\vspace{4mm}
  \includegraphics[width=\linewidth]{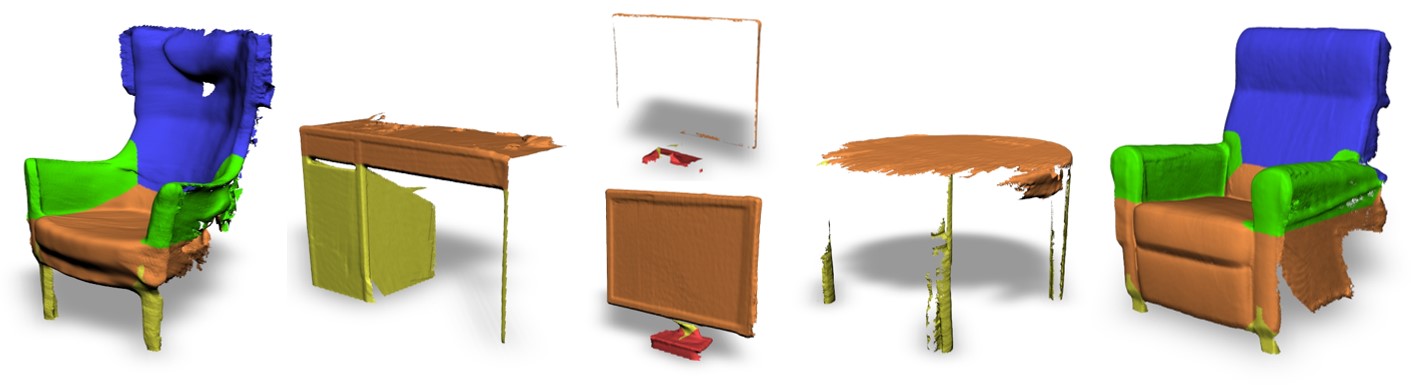}
  \caption{ \small Semantic labeling for partial point scans of shapes, using our pipeline for structure transfer.}
  \label{fig:partial_shape_labeling}
\end{figure}

\subsubsection{Fitting the known shape to the scan}
\label{subsubsec:skinning}

Our templates provide an intuitive way of simultaneously manipulating all shapes sharing the template structure: in this case, we can deform the collection shape $S_{\text{source}}$ to match the partial scan $S_{\text{partial}}$. This will recover a complete shape matching the scan. Since finding $S_{\text{source}}$ takes into account the template fit, the amount of distortion that this deformation process induces to $S_{\text{source}}$ is typically minimal.

\paragraph{Template-based Deformation} Assume we are given two shapes $S_1$, $S_2$ with compatible structure; namely, their best-fitting templates (Section~\ref{sec:template_selection}) have the same graph structure, but different parameter values. We denote the two parameterized fitted templates for the two models by $B_1$ and $B_2$, both with $n$ boxes. The goal now is to ``morph'' $S_1$ to $S_2$. Our simple approach is inspired by traditional skinning techniques \cite{sederberg1986_free-form-deformation-of-solid-geometric,joshi2007_harmonic-coordinates-for-character-articulation,ju2005_mean-value-coordinates-for-closed}: since there is a box-to-box correspondence between $B_1$ and $B_2$, we define the transformation of $S_1$ to $S_2$ via a weighted sum of the affine transformations $\{\mathcal A_i(.)\}$  that map each individual box $B_{1,i}$ to its corresponding box $B_{2,i}$, for $i = 1, 2, ... ,n$. The parameters of $B_{1,i}$ are its center  $\mathbf c_1 = [c_1^x,c_1^y,c_1^z]^T$  and its dimensions $\mathbf l_1 = [l_1^x,l_1^y,l_1^z]^T$ -- similarly for $B_{2,i}$. Then, the affine transformation mapping each point $\mathbf p_1$ on $B_{1,i}$ to its corresponding point $\mathbf p_2$ on $B_{2,i}$ is given by
 % \begin{equation}
 $
   \mathbf p_2 = \mathbf c_2 + \mathbf R_{12} (\mathbf p_1 - \mathbf c_1)
   % \label{equ:box_transformation}
 % \end{equation}
$
where $\mathbf R_{12} = \textrm{diag}\left(\frac{l_2^x}{l_1^x}, \frac{l_2^y}{l_1^y}, \frac{l_2^z}{l_1^z}\right)$.
Using the individual box transformations, any given point $\mathbf q_1$ on $S_1$  is mapped to a point $\mathbf q_2$ on $S_2$ via: $ \mathbf q_2 = \mathcal{A}(\mathbf q_1) = \frac{\sum_i w_i(\mathbf q_1) \mathcal A_i(\mathbf q_1)}{\sum_i w_i(\mathbf q_1) }$.
%\begin{equation}
%  \mathbf q_2 = \mathcal{A}(\mathbf q_1) = \frac{\sum_i w_i(\mathbf q_1) \mathcal A_i(\mathbf q_1)}{\sum_i w_i(\mathbf q_1) }
%  \label{equ:skinning}
%\end{equation}
This yields a continuous full shape transformation. We choose the weights to be in inverse relation to the distance between the point and the boxes, to ensure that any one box only locally affects the points inside and around it. Additionally, we want the weights to be smooth, to preserve smoothness of the underlying geometry. We use  $w_i(\mathbf q) = \exp\left(-d(\mathbf q,i)^2\right)$, where $d(\mathbf q,i)$ denotes the distance between point $\mathbf q$ and the $i$-th box in the template.
%
%
% Consider a random point $p$, at distances $(d_{p,1},d_{p,2},...,d_{p,n})$ from the $n$ boxes. The points each of the box transformations take point $p$ to  are given by $(L_1(p), L_2(p), ... , L_n(p))$. If any of the values $d_{p,i} = 0$, we would need the point $p$ to be transformed to $L_i(p)$, and in any other case the point $p$ needs to be transformed to a combination of the points given by $L_j(p)$ for $j = 1,2,...,n$. A naive way to perform this transformation is to assign the point $p$ to the box $B_{1,i}$, to which it is closest, i.e. $d_{p,i} = \text{min}_j d_{p,j}$ if $p$ is assigned to box $B_{1,i}$. Each point $p$ would mirror the transformation of the box it is assigned to. But in this method, points close to two adjacent boxes would be weighed by one of the boxes but not the other, and therefore the skewing close to such junctions would be very bad.
%
%
After mapping  all points on $S_1$ using the process above, we perform a global scaling so that the final deformed shape $\tilde S_1$ lies in the same global bounding box as $B_2$ and thus matches its proportions, but preserves the shape details from $S_1$.
%
% Note that since we don't uniquely assign a point to a single box, but instead use a weighted transform, we cannot guarantee that points lying \emph{on} a certain box will  be lying on the matching box after the transformation. This is acceptable in principle, since the boxes are intended to capture structure more than geometry: even on an un-transformed object, it is not necessary that the entirety of a structural part lies entirely inside a box, it is merely assigned to it. We found that the continuity advantage of a weighted, skinning-like transform behaves very well in practice.
%
Note that $S_1$ and $S_2$ can be represented by either meshes or point clouds at this stage. In the former case, we transform the vertices of $S_1$ and keep the connectivity the same.

% Since this shape adjustment is performed on shapes very close in parameterization to the original shape and the weights are fairly sluggish with increasing distance from the boxes, this adjustment is observed to be fairly compact on input shapes.

\paragraph{Fitting source to partial} In order to recover the best fitting shape  for our partial scan, we simply apply the process above with $S_1 = S_{\text{source}}$ and $S_2 = S_{\text{partial}}$. The output $\tilde S_{\text{source}}$ is the recovered model.

\begin{table}
  \scalebox{0.9}{
\begin{tabular}{c|c|m{1.5cm}|m{2.0cm}|m{1.0cm}}
  \hline
 { \bf Category} & {\bf \#Shapes} & {\bf PointNet} & {\bf SyncSpecCNN} & {\bf Ours} \\
  \hline
Chair & 3746  & 89.6 & {\bf 90.24} & 87.37  \\
Table & 4520 & 80.6  & 82.13  & {\bf 88.62}\\
Cup & 184  & 93.0 & 92.73 & {\bf 94.06} \\
Lamp & 1547  & 80.8 & {\bf 84.65} & 78.03\\
Airplane & 2690 & {\bf 83.4} & 81.55 & 76.71 \\
\hline
\end{tabular}
}
\caption{ \small Intersection-over-Union (IoU) percentages for our part labeling, evaluated against the ground truth and compared to \protect\cite{DBLP:journals/corr/QiSMG16} and \protect\cite{yi2016syncspeccnn}. The highest IoU value for each category is in bold. }
\label{tab:partlabel}
% \vspace{-8mm}
\end{table}
\begin{table}
\centering
  \scalebox{0.9}{
\begin{tabular}{l|c|c|c}
  \hline
 \multirow{2}{*}{\bf Task} & \multicolumn{3}{c}{\bf Accuracy} \\
\cline{2-4}
& Best & Best 2 & Best 3 \\ \cline{2-4}
  \hline
Cluster Classification - Training & 89\% & 94\% & 98\% \\
Cluster Classification - Testing & 73 \% & 83 \% & 92 \% \\
\hline
\end{tabular}
}
\caption{ \small Performance of the dual-classification network for template and cluster classification on partial virtual scans of ShapeNet objects. The three columns report the percentage of cases where the correct fit (cluster or template) was found among the 1, 2 or 3 most likely fits according to the net's output.}
\label{tab:nets}
\end{table}

\begin{figure}
  \includegraphics[width=\linewidth]{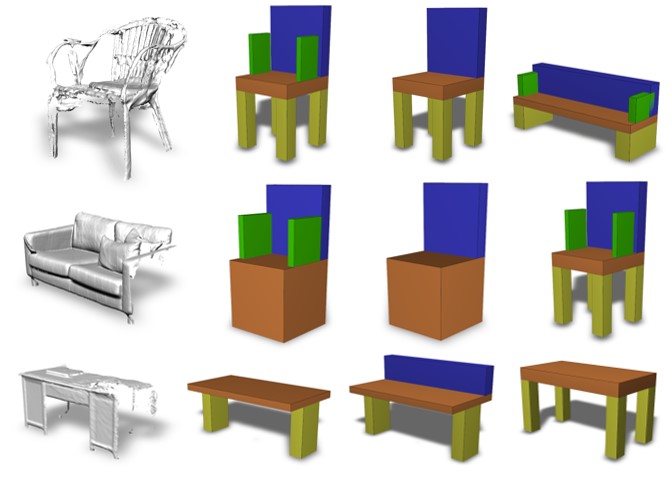}
  \caption{ \small The three most likely templates, as determined by the neural network for given partial scans. Left to right: input partial scan and possible structure templates (pre-parameter fitting), in decreasing order of likelihood.}
  \vspace{-4mm}
  \label{fig:best_template_nn_result}
\end{figure}

%!TEX root = ../egpaper_submission.tex

\section{Results and Discussion}

The pipeline has been implemented in C++ -- we plan to make the code freely available online. Our shape collection is a subset of the ShapeNet repository \cite{chang2015_shapenet:-an-information-rich-3d-model-repository}. We used shapes from 10 different categories: monitors, cups, tables, chairs, benches, desks, dressers, sofas, airplanes, and lamps.

\subsection{Template Fitting and Selection for Database Shapes}
We fit templates running the CMA-ES \cite{hansen2006cma} algorithm for four different initializations. Some examples of fitting template parameters to shapes in the collection $\mathcal S$ (Section~\ref{sec:collection_structure}) are shown in Fig.~\ref{fig:template_correspondences}. The success of this process is critical for various other results, eg. shape manipulation (Fig.~\ref{fig:skinning}) and part labeling (Fig.~\ref{fig:shape_labeling}), and thus fitting results are implicitly showcased as an intermediate step in these figures.

\subsection{Template-Based Deformation}
To evaluate our template-based deformation (Section~\ref{subsubsec:skinning}), we show some results with complete shapes in Fig. \ref{fig:skinning}. We show the source $S_1$  and target $\tilde S_1$ shapes, their fitted templates, and the deformed version of $S_1$ that aligns to $\tilde S_1$. Since all shapes are scaled to fit the figure, we measure the magnitude of induced deformation by via the normalized Euclidean distance between each vertex in $S_1$ and its mapped location in $\tilde S_1$, and show it as a heat map. The box-based transformations help individually adapt, and globally align, the individual semantic parts, without drastically losing their individuality (while the overall sizes of seats in Fig.~\ref{fig:skinning} match, they do not deform to align with one another). This enables generating variations of shapes in a structure-preserving way, which could be interesting  for editing or morphing applications; this is however not a focus of this paper.

\begin{figure*}[t]
  \includegraphics[width=\linewidth]{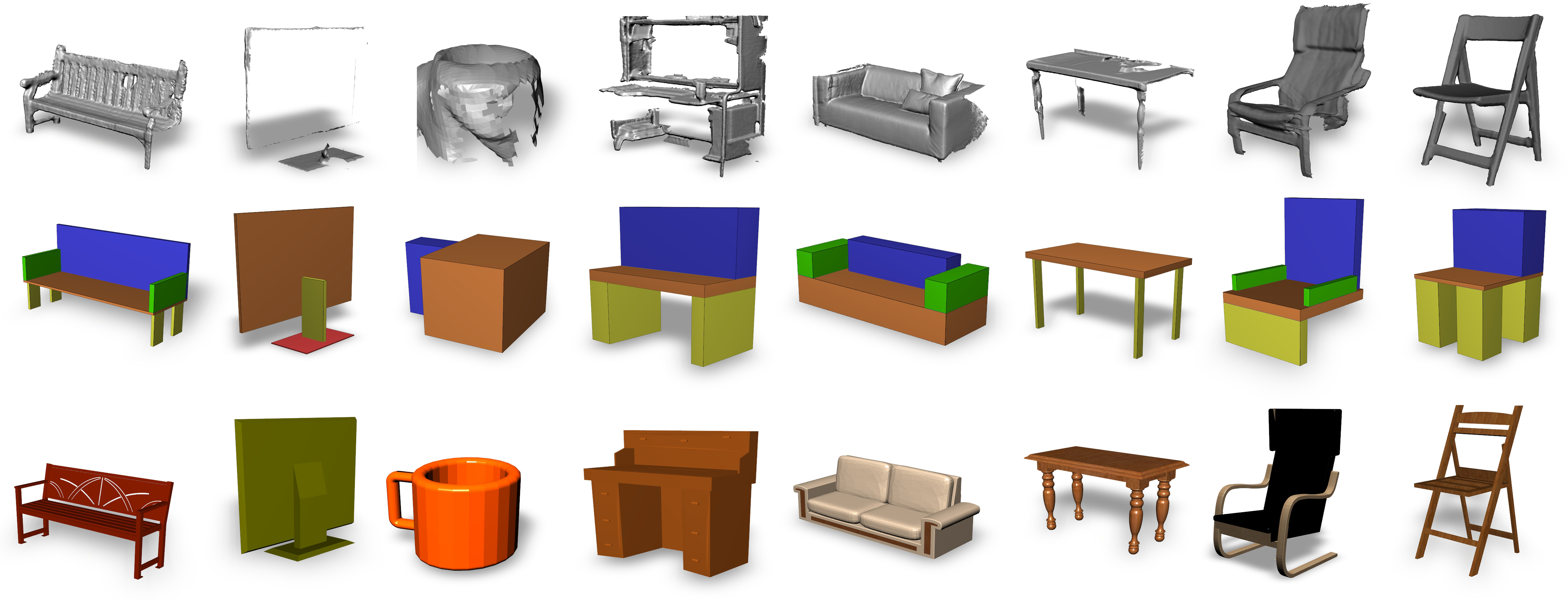}
  \caption{ \small Partial shape recovery on the real-data reconstruction dataset~\protect\cite{DBLP:journals/corr/QiSNDYG16}, for 4 different categories, bench, monitor, mug, desk. Top row: partial scans of entirely new objects (i.e. not present in the shape collection). Middle: optimization-based template fitting to the partial scan, after classification to template structure and geometric cluster. Bottom: recovered shapes from the collection, deformed to fit the partial scan.}
  \vspace{-4mm}
  \label{fig:partial_scan_reconstruction}
\end{figure*}

\subsection{Deep Network Training and Output}
Table \ref{tab:nets} shows the accuracy of cluster classification achieved by the classification network for partial scans of shapes. The network operates on $22051$ signed distance fields obtained from virtually generated partial scans of shapes from the ShapeNet repository, across the $8$ different shape categories. We used 80\% of this data for training and the remaining for testing. As is also evident in Table \ref{tab:nets}, using the net to select the \emph{single} best cluster for a partially scanned shape may be a suboptimal strategy, since the classification accuracy is not always high. Our strategy of exploring the best three cluster predictions ensures that we are operating at very high classification accuracy. This, in return, ensures that the template fitting is initialized with a much better estimate. The choice of a number of top clusters thus provides a trade-off between quality of initialization and post-processing time needed to perform the optimization rounds. %The role of selected clusters in making the template fitting optimization faster is discussed further in the supplementary material\todo{is it?}.

Qualitative results of the network output on new partial shape scans are shown in Fig.~\ref{fig:best_template_nn_result}. The figure shows the most likely templates, as established by the clusters, which at this stage encode the structure and also provide a rough initialization for the template fitting. (Section~\ref{subsec:partial_shape_recovery}).

% As mentioned in Section~\ref{subsec:partial_shape_recovery_completion}, employing the network ensures performing costly fitting optimizations between all possible templates and the partial scan.

% % The role of the network is crucial in two aspects. If the best template and cluster pair is not known, it would become mandatory to test the partial scan against every template and every cluster initialization pair. Since it is known from Table \ref{tab:cmaes} that the optimization is time-intensive, and could take upto 10s for convergence per each template-cluster initialization pair on average, this time is reduced to effectively at most $90$s of post-processing once the network provides us with the dual-classification information. This is the first advantage of using the network.
% The actual time taken to perform post-processing could vary depending on how many top pairs of clusters and templates from the network are chosen to be initialized with. This provides a trade-off between quality of initialization and the time taken to perform optimization.

\subsection{Partial Shape Recovery}

Recovering shapes from partial point scans, using the network output and the identification/fitting process of Sections~\ref{subsubsec:partial_shape_identification} and \ref{subsec:partial_shape_recovery} is shown in Fig.~\ref{fig:partial_scan_reconstruction}.
% Note that our shape recovery pipeline produces complete CAD models as output, since we are retrieving them from the shape database, which (as opposed to raw point clouds) can, for example, directly be used for modeling purposes, which can be useful for AR or CG applications.
% In addition, our pipeline goes beyond simple shape retrieval, in that it also automatically deforms the retrieved shape to fit the detected structure of the partial point scan.
%
Our template fitting optimization is robust to partiality as well as noise in the point scans. This is highlighted in more detail in the supplementary material. As can be observed in the last two columns of Fig.~\ref{fig:partial_scan_reconstruction}, partial shape recovery on shapes that do not fit the classical box-template structure can be done effectively. In case of the folding chair, the dimensions of the template fit, with a thick backrest and thick legs, place the partial scan into a cluster containing other folding chairs that have the same dimensional properties on the back and legs. Similarly, though the fit of the S-shaped  partial scan does not contain the legs, the dimensions of the remaining parts aid in recovering a similarly  shaped  object from the database to complete the partial scan.
In addition to reconstructing a point cloud by recovering a complete mesh from the collection, we can also utilize the inferred structure to augment the point cloud itself. Since this is not the main focus of the paper, we provide some point cloud completion techniques in the supplementary material with some qualitative results.

\begin{figure}
  \includegraphics[width=\linewidth]{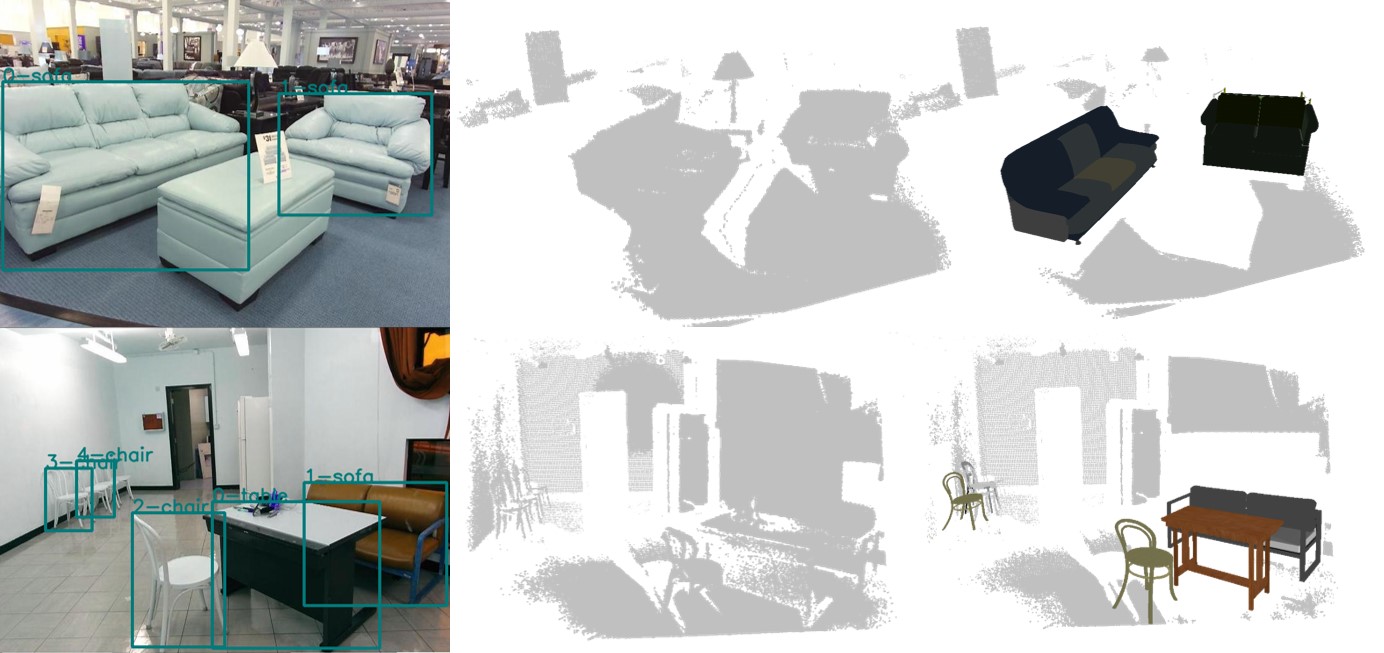}
  \caption{ \small Scene completion with full mesh replacements for identified objects in a partially generated scene. Left to right: RGB images with bounding box annotations on objects, point cloud generated using ~\protect\cite{DBLP:journals/corr/abs-1711-08488}, scene completion provided by our fitting method on the bounding boxes.}
  \vspace{-4mm}
  \label{fig:scene_completion}
\end{figure}

\paragraph{Scene completion.}
In Fig.~\ref{fig:scene_completion}, we provide an example of recovering shapes for real-life RGBD scanned indoor scenes. We preprocess the scenes using ~\cite{DBLP:journals/corr/abs-1711-08488}, which detects 3D objects in the scene and annotates the RGBD point cloud with the objects' 3D bounding boxes. We treat the points in each of these bounding boxes as a partial scan, recover a fitted CAD mesh using our structure-aware recovery pipeline, and replace scanned points with the recovered mesh.
% We show two examples of this in the form of indoor scenes, the first containing two annotated sofas, and the second containing one sofa, one table and three chairs.
The retrieved shapes are fairly close to the input shapes despite heavily intersecting bounding boxes. Replacing scanned points in the bounding boxes with retrieved shapes makes the scene less cluttered and would allow for further scene editing. Note that the failure to recover lamp and sofa meshes in the first scene is due to their bounding boxes not being detected by the aforementioned method.
% In case the annotations are incomplete, though, especially in cases where the missed out objects lie in one or more of the bounding boxes of other objects, these points are completely lost in the completion. This is seen in the case of the lamp and the sofa table in the first scene, both of which are not annotated in the image, and hence, miss out on the completed scene.

% We do not perform quantitative comparisons to either of these methods because both and the goals and the output format of the three methods are all different.

\subsection{Comparisons.}
\begin{figure}
  \centering
  \includegraphics[width=\linewidth]{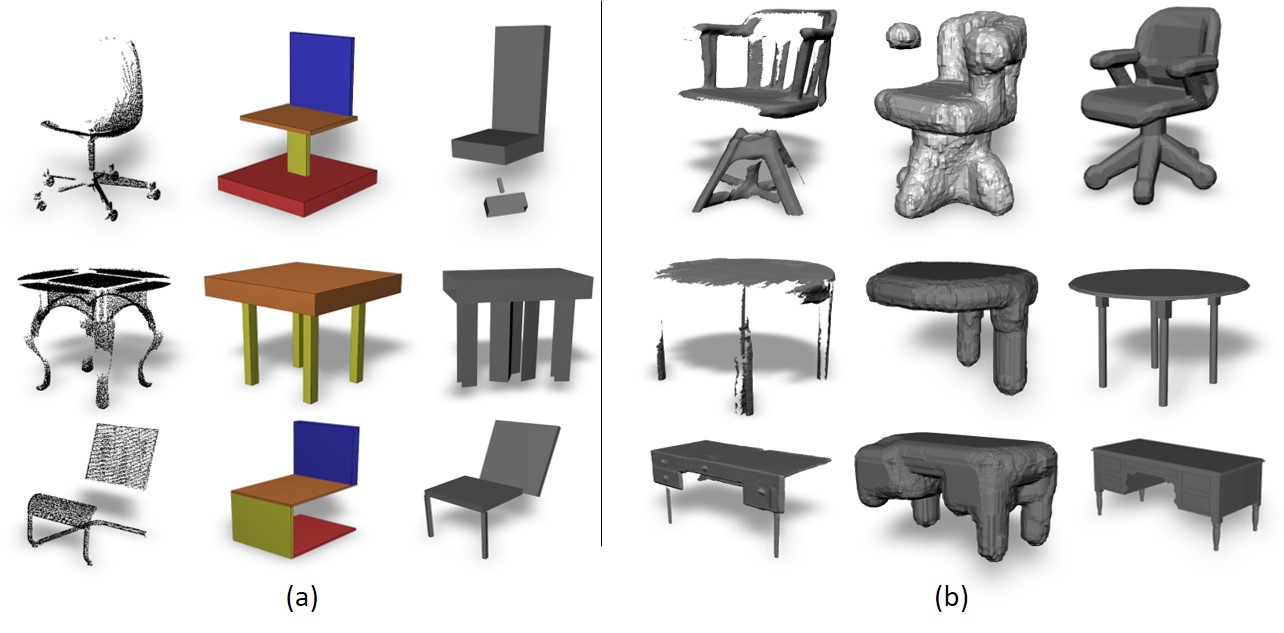}
  \caption{ \small (a) Comparison to Sung et al.\protect\shortcite{Sung:2015:DSP:2816795.2818094} Columns, left to right: partial point clouds, our templates fit to the input, box fits generated by Sung et al. to the input. (b) Comparison to Dai et al. Columns, left to right: input partial scans, shape completion obtained by applying the end-to-end technique proposed in Dai et al., our shape retrieval and fitting method applied on the input.}
  \vspace{-4mm}
  \label{fig:comparison_sung_dai}
\end{figure}

In Fig.~\ref{fig:comparison_sung_dai}(a), we qualitatively compare against the work of Sung et al. \shortcite{Sung:2015:DSP:2816795.2818094}, which also retrieves box-based parts for an input partial point cloud. In their case, parts are retrieved individually via  optimization -- in comparison, our method tends to provide more realistic part layouts due to the structure enforced by the templates. Fig.~\ref{fig:comparison_sung_dai}(b) provides a qualitative comparison to the end-to-end technique of Dai et al.~\shortcite{dai2016complete}. Since no structure is available, this technique provides somewhat ``blobby'' completions ; while these results could  be used for object classification tasks or as training datasets, the recovered shape is too rough to be used as as a prototype e.g. in AR/CG scene editing applications. In contrast, our pipeline provides a full CAD mesh, fitted to match the input scan, which can  be directly be used in place of the partially scanned shape e.g. for scene modeling purposes.

In the inset figure, we show a comparison to the fit produced by Tulsiani et al. ~\shortcite{DBLP:journals/corr/TulsianiSGEM16}. Even though their result
\begin{wrapfigure}{r}{0.3\textwidth}
\vspace{-10pt}
\hspace{-10pt}
  \centering
  \includegraphics[width=\linewidth]{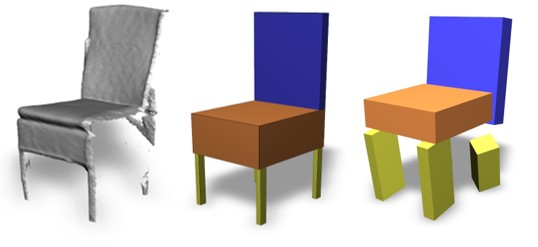}
  %\caption{ \small Ours (middle) vs. Tulsiani et al.'s ~\protect\shortcite{DBLP:journals/corr/TulsianiSGEM16} (right) fitting to an input partial point cloud (left).}
  % \hspace{-20pt}
\vspace{-20pt}
\end{wrapfigure}
is regressed by
a network trained towards box-based fitting of chairs specifically, the lack of a clearly defined structure among the boxes makes it difficult to correctly fit to the shape. In comparison, our constrained templates  recover the missing parts of the shape more accurately. For fairness, we note that Tulsiani et al. \shortcite{DBLP:journals/corr/TulsianiSGEM16} do not aim to complete partial shapes.

\subsection{Semantic Part Labeling}
The box-primitives in our shape templates are, by design, associated with real shape parts with a ``semantic'' meaning -- e.g. legs or armrests in a chair template. Fitting a template transfers this  information onto a shape and annotates it with meaningful parts.  We annotate each point on a source shape $S$ by assigning the index of the box closest to it (by projection distance) in its best-fitting template $T^*(S)$. This can be done both for partial scans (Fig.~\ref{fig:partial_shape_labeling}) and for complete shapes, e.g. from the shape collection (Fig.~\ref{fig:shape_labeling}). %Note that our primitive-based approach is tailored to man-made shapes, but is still applicable to organic shapes.  Airplanes, chairs, lamps, and mugs are from ShapeNet \cite{chang2015_shapenet:-an-information-rich-3d-model-repository}. Animals are from TOSCA \cite{bronstein2008_numerical-geometry-of-non-rigid-shapes}; we only used animals in an identical, non-deformed canonical pose, so that their shape can roughly be described with primitives.

Table \ref{tab:partlabel} discusses accuracy of part labeling on 5 categories. We consider the labeling by \cite{yi2016scalable} as ground-truth, where shape part labels are obtained in a semi-supervised way, using Mechanical Turk verification. We evaluate our technique against the performance of \cite{DBLP:journals/corr/QiSMG16} and \cite{yi2016syncspeccnn} on the same task. While both these methods use supervision on $70 \%$ of the category size in ShapeNet, with an additional $10\%$ of the shapes used in the validation set,
  % - onl being  and testing set size being $20\%$ of the overall category size.
we use no external supervision to perform the part labeling.

% \begin{figure}
%   \includegraphics[width=\linewidth]{images/energy.pdf}
%   \caption{ \small Comparing various modes of initialization for the optimization of partial scan. This plot shows the comparison of fitting a template to a partial shape using the optimization based on random (blue), standard (red) (as in Section \ref{sec:optimization}) and neural network predicted (green) optimizations. The data plotted is the logarithm of the average energy functional  against iterations. The random initialization on average converges to a local minimum, while both the standard initialization and the network-predicted initialization converge to the global optimum. It is also seen that the standard initialization (red curve) takes around 200 iterations on average to converge, while the network-predicted initialization converges within 75 iterations on average. This shows that the neural network speeds up our partial scan recovery process.  \todo{cite me}}
%   \label{fig:effect_of_initialization}
% \end{figure}

%!TEX root = ../egpaper_submission.tex
\section{Conclusion and Future Work}

Obtaining structural information about an object, scanned by commodity hardware into an unstructured partial point cloud, can be key to identifying the object and reasoning about its functionality.
%In this paper, we present a pipeline for leveraging the contents of a shape collection to transfer structural information onto new, partially scanned shapes.
We represent structure by a set of pre-designed structural templates, based on simple box-like primitives.
%We defer the difficult problem of structure inference to a neural network, which provides us with the most likely structural pattern for the partial scan based on the contents of a shape collection.
We leverage the obtained structural information using a neural network, and show applications of recovering the shape of a partial scan, annotating its structural parts, and applying this to perform scene completion.
We provide a single lightweight pipeline that achieves good performance in all these tasks.
Our method is unique in that it recovers a full mesh to account for a partial scan of an object.
This highlights the value of simple hand-crafted templates, which can abstract away significant geometric detail. That said, automatically inferring the shape templates, and even incorporating different primitives, is the ideal scenario; it remains  a difficult unsolved problem, especially when both template parts and inter-part symmetries need to be inferred providing an interesting avenue for future work.

\subsection*{Acknowledgements}
The authors would like to acknowledge NSF grants IIS-1528025, DMS-1546206, a Google Focused Research Award, a Google Research Award, gifts from the Adobe Corporation, Amazon, Google, and the NVIDIA Corporation, a Swiss National Foundation Early PostDoc. Mobility Grant P2EZP2-165215 and a Stanford AI Lab-Toyota Center for Artificial Intelligence Research grant.

{\small
\bibliographystyle{ieee}
\bibliography{egpaper_submission.bbl}
}

\end{document}